\documentclass[acmtog,nonacm,numbers]{acmart}

\renewcommand\footnotetextcopyrightpermission[1]{} 
\usepackage{booktabs} 
\usepackage{multirow} 
\usepackage{xcolor} 
\usepackage{caption}
\usepackage{cleveref}
\usepackage{pifont} 
\usepackage{textcomp} 
\usepackage{xspace}
\usepackage{makecell}
\usepackage{enumitem}
\usepackage{hyperref}

\usepackage[title,titletoc]{appendix}
\usepackage{placeins}

\usepackage[ruled]{algorithm2e} 

\SetAlFnt{\small}
\SetAlCapFnt{\small}
\SetAlCapNameFnt{\small}
\SetAlCapHSkip{0pt}

\crefname{table}{Table}{Tables}
\crefname{figure}{Figure}{Figures}
\crefname{equation}{Equation}{Equations}
\crefname{section}{Section}{Sections}

\acmJournal{TOG}


\definecolor{lightgray}{RGB}{200, 200, 200}

\newcommand{\method}{MVP4D\xspace}

\author{Felix Taubner}
\orcid{0000-0002-9717-7181}
\email{ftaubner@cs.toronto.edu}
\affiliation{%
    \institution{University of Toronto and Vector Institute}
    \country{Canada}
}
\author{Ruihang Zhang}
\orcid{0009-0004-8989-4377}
\email{ruihang.zhang@mail.utoronto.ca}
\affiliation{%
    \institution{University of Toronto}
    \country{Canada}
}
\author{Mathieu Tuli}
\orcid{0009-0000-9112-0048}
\email{tuli.mathieu@gmail.com}
\affiliation{%
    \institution{LG Electronics}
    \country{Canada}
}
\author{Sherwin Bahmani}
\email{bahmani@cs.toronto.edu}
\orcid{0009-0009-5724-3402}
\affiliation{%
    \institution{University of Toronto and Vector Institute}
    \country{Canada}
}
\author{David B. Lindell}
\orcid{0000-0002-6999-3958}
\email{lindell@cs.toronto.edu}
\affiliation{%
    \institution{University of Toronto and Vector Institute}
    \country{Canada}
}

\begin{document}

\title{MVP4D: Multi-View Portrait Video Diffusion for Animatable 4D Avatars}

\begin{teaserfigure}
  \includegraphics[width=\textwidth]{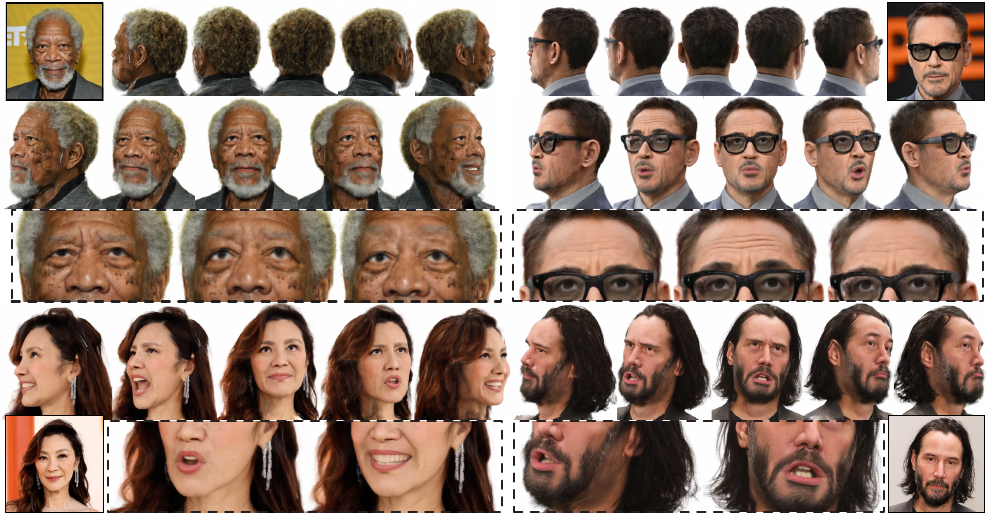}
  \vspace{-2.5em}
  \caption{MVP4D generates animatable and photorealistic 360-degree avatars from a single reference image using a multi-view video diffusion model. The model generates videos of the reference subject from multiple viewpoints with controllable expressions and head pose, which we distill into a 4D avatar that can be rendered in real time.  Our method produces detailed 4D avatars with realistic motion, capturing fine features such as skin blemishes, hair strands, and expression-dependent wrinkles. \textbf{\emph{Please refer to the project webpage for video results and code: \href{https://felixtaubner.github.io/mvp4d/}{\textcolor{blue}{\texttt{https://felixtaubner.github.io/mvp4d/}}}.}}}  
  \label{fig:teaser}
\end{teaserfigure}

\begin{abstract}
Digital human avatars aim to simulate the dynamic appearance of humans in virtual environments, enabling immersive experiences across gaming, film, virtual reality, and more. However, the conventional process for creating and animating photorealistic human avatars is expensive and time-consuming, requiring large camera capture rigs and significant manual effort from professional 3D artists. With the advent of capable image and video generation models, recent methods enable automatic rendering of realistic animated avatars from a single casually captured reference image of a target subject. While these techniques significantly lower barriers to avatar creation and offer compelling realism, they lack constraints provided by multi-view information or an explicit 3D representation. So, image quality and realism degrade when rendered from viewpoints that deviate strongly from the reference image. Here, we build a video model that generates animatable multi-view videos of digital humans based on a single reference image and target expressions. Our model, MVP4D, is based on a state-of-the-art pre-trained video diffusion model and generates hundreds of frames simultaneously from viewpoints varying by up to 360 degrees around a target subject. We show how to distill the outputs of this model into a 4D avatar that can be rendered in real-time. Our approach significantly improves the realism, temporal consistency, and 3D consistency of generated avatars compared to previous methods.
\end{abstract}
\maketitle

\section{Introduction}

Digital human avatars aim to simulate human appearance, behavior, and interaction in virtual environments.
Their development enables immersive experiences across applications, including gaming, film, virtual reality, and education.
However, creating and animating realistic avatars remains challenging due to the difficulty of (1) capturing and faithfully reproducing the subtleties of human appearance, motion, and expression, and (2) human sensitivity to even minor rendering imperfections, which often leads to the “uncanny valley” effect~\cite{geller2008overcoming,mori1970uncanny}.

Conventionally, 4D avatars are created using large capture rigs containing many cameras~\cite{debevec2012light,he2024diffrelight,wuu2022multiface}, and creating the final avatar demands significant manual effort from professional 3D artists. 
Animation often involves complex motion capture setups and keyframing, which is expensive and time-consuming~\cite{unreal2024thewell,haapaoja2022advances,seol2016creating}.
Hence, recent research in computer graphics focuses on automatically creating and animating photorealistic avatars from casually captured images or videos of a target subject using a single camera \cite{taubner2024cap4d,qian2024gaussianavatars,xiang2024flashavatar,zielonka2022insta,shao2024human4dit,ma2024followyouremoji}.

Given one or more reference images, recent approaches reconstruct and render avatars using either 3D representations---such as Gaussian splatting~\cite{kerbl20233d,hu2024gaussianavatar} or volume rendering~\cite{xu2024vasa,drobyshev2022megaportraits}---or 2D generative techniques that directly synthesize avatar videos~\cite{ma2024followyouremoji,lin2025omnihuman,cui2024hallo3}.
These approaches face a fundamental tradeoff: 3D methods offer stronger multi-view consistency but often struggle to capture fine-grained facial dynamics and expression-dependent appearance due to modeling limitations~\cite{lewis2014practice,emoca}.
In contrast, 2D methods leverage large-scale generative models~\cite{rombach2022high,saharia2022photorealistic,brooks2024video,gao2024cat3d} that better capture subtle motion and expressions, but they lack explicit 3D constraints and hence exhibit weaker multi-view consistency.

Recent advances in generative modeling have focused on enhancing the 3D consistency of 2D methods by jointly modeling and rendering multi-view images~\cite{gao2024cat3d,wu2024cat4d,kant2025pippo,taubner2024cap4d} or multi-view videos~\cite{shao2024human4dit}.
Still, current approaches for multi-view human video generation rely on image-based diffusion models, which limits animation fidelity and constrains generation to only a few frames---hindering the creation of compelling, temporally coherent animations.

Here, we address these limitations by introducing \method---a technique based on a morphable multi-view video diffusion model (MMVDM) that generates detailed, photorealistic avatars from a single reference image (see Figure~\ref{fig:teaser}).
Our model is based on a recent video diffusion transformer architecture~\cite{yang2024cogvideox}, which we extend using morphable model-based control signals to enable high-fidelity animation and 3D-consistent rendering.
We produce 3D- and temporally consistent multi-view videos in a single diffusion sampling run, generating up to 400 frames across 360 degrees of viewpoint variation (see Table~\ref{tab:compare_generators}).
Based on these videos, we show how to distill the generated multi-view portrait videos into a 4D avatar that can be rendered in real time.

Training the MMVDM is challenging due to the scarcity of multi-view portrait video data and the computational requirements of training large video models.
To overcome this obstacle, we design a multi-modal training curriculum that combines the temporal coherence of single-view dynamic videos with the view consistency of multi-view static images.
At inference time, this enables the model to synthesize long, synchronized video sequences from multiple viewpoints---without ever being trained on large-scale multi-view video data.

In summary, we make the following contributions.
\begin{itemize}[topsep=0pt,leftmargin=*]
    \item We introduce the first morphable multi-view video diffusion model capable of synthesizing 360-degree portrait videos.
    \item We propose a multi-modal training curriculum that enables portrait synthesis across a large number of frames and viewpoints, without requiring explicit training on large-scale multi-view video data.
    \item We demonstrate state-of-the-art results in generating dynamic, temporally consistent, and photorealistic multi-view portrait videos and 4D avatars. 
\end{itemize}

\newcommand{\yes}{\textcolor{green!60!black}{\ding{51}}} 
\newcommand{\no}{\textcolor{red}{\ding{55}}}             

\begin{table}[t]
    \centering
    \caption{\textbf{Comparison of recent multi-view and video generation models.} \method generates an order of magnitude more \textbf{frames} in a single diffusion sampling run than comparable baselines, while producing temporally consistent \textbf{video} across \textbf{multiple viewpoints}. In addition, it is conditioned on head \textbf{pose} and \textbf{expression}, making it fully animatable.}
    \resizebox{\columnwidth}{!}{
    \begin{tabular}{ l|c c c c } 
        \toprule
        method & \#frames & multi-view & video & pose/exp. \\
        \midrule
        CAT3D~\cite{gao2024cat3d} & 8 & \yes & \no & \no \\
        CAT4D~\cite{wu2024cat4d} & 16 & \yes & \yes & \no \\
        CAP4D~\cite{taubner2024cap4d} & 8 & \yes & \no & \yes \\ 
        CogVideoX~\cite{yang2024cogvideox} & 49 & \no & \yes & \no \\
        Pippo~\cite{kant2025pippo} & 60 & \yes & \no & \no \\ 
        Human4DiT~\cite{shao2024human4dit} & 24 & \yes & \yes & \yes \\ 
        \midrule
        MVP4D & 392 & \yes & \yes & \yes \\ 
        \bottomrule
    \end{tabular}}
    \label{tab:compare_generators} 
     \Description{This table shows a comparison of the MVP4D MMVDM to other multi-view generation models in terms of number of frames, multi-view, video and pose/expression control. }
\end{table}

\section{Related Work}

\paragraph{Animatable 3D avatars.}
Recent methods for animating and rendering digital human avatars typically rely on 3D representations of appearance or geometry.
For example, techniques such as neural radiance fields \cite{mildenhall2021nerf} and Gaussian splatting \cite{kerbl20233d} use models of the absorption and emission of light in a 3D volume to render photorealistic images of scenes \cite{zielonka2022insta} or digital humans \cite{xu2024gaussian,hu2024gaussianavatar,shao2024splattingavatar,xiang2024flashavatar}.
However, these approaches are typically applied to the reconstruction problem: given a large set of images captured of a single human from multiple view directions, they reconstruct appearance and geometry.
Other approaches based on 3D representations focus on the task of generation---by training on a large corpus of videos of humans, they learn to generate animatable 3D human models from one or few input images of a person, e.g., through feed-forward generation \cite{xu2024vasa,drobyshev2022megaportraits,voodoo3d,khakhulin2022realistic,chu2024gagavatar, zheng2025headgap,tran2024voodooxp} or by inverting 3D generative adversarial networks~\cite{sun2023next3d,zhao2024invertavatar}.
More recent methods, such as Taubner et al.\shortcite{taubner2024cap4d}, generate many additional reference images using a morphable multi-view diffusion model (MMDM) to improve the 3D reconstruction.
Still, animating these 3D representations is challenging. Most methods attempt to model motion via coarse geometry generated by 3DMMs, which fails to capture the fine-grained motion of structures such as the tongue and lips.

\begin{figure*}[ht]
\includegraphics[width=\textwidth]{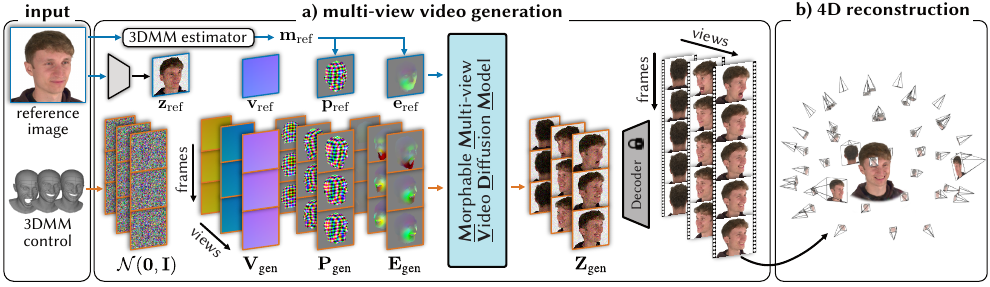}
\caption{\textbf{Overview of MVP4D.}  The method takes as input one reference image that is encoded into the latent space of a variational autoencoder~\cite{yang2024cogvideox}. An off-the-shelf face tracker estimates a 3DMM for the reference image, from which we derive conditioning signals that describe camera pose, head pose, and expression. We associate additional conditioning signals with each input noisy latent video frame based on the desired generated viewpoints, poses, and expressions. A multi-view video diffusion transformer denoises the latent multi-view video. We then use the variational autoencoder to decode the latent video frames to a series of multi-view image frames (the resulting multi-view video), which is distilled into a 4D representation for real-time visualization.}
\label{fig:method} 
\Description{This figure shows an overview of the MVP4D pipeline. }
\end{figure*}

\paragraph{Monocular portrait video generation.}
Another line of research focuses on directly generating videos of the digital human avatars from a reference image without relying on an intermediate 3D representation.
These approaches use image or video generation models trained on large-scale datasets \cite{rombach2022high,blattmann2023stable} to output animated avatar videos from an input image of a person \cite{tian2024emo, wei2024aniportrait, yang2024megactor, lin2025omnihuman, ma2024followyouremoji,zhao2025xnemo, xu2025hunyuanportrait}.
They are typically controlled using the facial expressions of an example video or from an input audio clip.
These methods show impressive results; by learning directly from large-scale monocular video collections of humans, they achieve near-photoreal quality and lifelike motion.
However, they do not provide a 3D representation and often exhibit artifacts when the rendered viewpoint deviates significantly from the input image, due to a lack of constraints that enforce 3D consistency.
Also, while some methods can use sparse facial landmarks to control the appearance of the generated avatar, there is no option for fine-grained controllability of the exact expression or head pose~\cite{guo2024liveportrait, siarohin2019first, ma2024followyouremoji}. 

\paragraph{Multi-view generation models.}
Recently, progress in diffusion models has enabled the generation of multi-view images or videos based on a single reference image~\cite{gao2024cat3d, wu2024cat4d}.
While earlier work~\cite{zhang20244diffusion,li2024vivid, jiang2024animate3d, liang2024diffusion4d} primarily focuses on simple object generation, more recent approaches have been applied to digital humans \cite{kant2025pippo,shao2024human4dit,taubner2024cap4d}.
Most similar to our method, Human4DiT \cite{shao2024human4dit} can generate body-pose controlled multi-view video given a reference image of a target likeness.
However, these methods are based on image generation models and generate only a very small number of frames at a time, thus greatly limiting temporal and 3D consistency.
With the recent advances in video diffusion models, several works extend pre-trained video models to multi-view video generators for general scenes~\cite{watson2024controlling, wang20244real,kuang2024collaborative,xu2024cavia}, driving simulation~\cite{li2024drivingdiffusion}, or panoramic scenes~\cite{xie2025videopanda}.
While these approaches demonstrate impressive quality, their application to human avatar generation remains unexplored.
Our model leverages recent video diffusion transformers \cite{yang2024cogvideox} with spatio-temporal autoencoders to generate a very large number of temporally and multi-view consistent frames.
See \cref{tab:compare_generators} for a comparison of multi-view generation models.
\section{Method}
\subsection{Morphable Multi-view Video Diffusion Model.}

\method employs a morphable multi-view video diffusion model (MMVDM) that takes a reference image $\mathbf{i}_\text{ref}$ as input. 
It generates a multi-view video $\mathbf{V}_\text{gen} = \{\mathbf{i}_\text{gen}^{(v,t)}\}_{v=1,f=1}^{V,F}$, i.e., a set of $G=V \times F$ images, where $V$ is the number of generated views and $F$ is the number of generated frames per view (see \cref{fig:method}).
Additionally, the model is conditioned on head pose, expression, and camera parameters for each image. 
These conditioning signals are denoted as $\mathbf{c}_\text{ref}$ for the reference image and $\mathbf{C}_\text{gen} = \{\mathbf{c}_\text{gen}^{(v,f)}\}_{v=1,f=1}^{V,F}$ for the generated images.
Based on these inputs, the MMVDM learns the joint probability of generated multi-view videos as
\begin{equation}
P(\mathbf{V}_\text{gen} | \mathbf{i}_\text{ref}, \mathbf{c}_\text{ref}, \mathbf{C}_\text{gen}). 
\label{eqn:mmvdm}
\end{equation} 

\subsection{Architecture}
\label{sec:architecture}
Our model is initialized from CogVideoX-2B~\cite{yang2024cogvideox}, and we adapt the architecture for multi-view generation similar to previous work~\cite{taubner2024cap4d,gao2024cat3d}.
Specifically, we use a pre-trained spatio-temporal video auto-encoder~\cite{yang2024cogvideox} to encode the reference image and each video from each view into a latent space with compressed spatial and temporal resolution.
The auto-encoder converts a monocular video (or image) of size ($[F=1+4\tilde{F}] \times H \times W$) to a $([1+\tilde{F}]\times \frac{H}{8} \times \frac{W}{8})$ latent video (or image). 
We introduce $\tilde{F}$ to account for the temporal compression scheme of the auto-encoder, which encodes the first input frame into its own latent frame and then applies a compression factor of four to all subsequent frames.

We encode the reference image, and each view of the multi-view video separately to obtain a reference latent $\mathbf{z}_\text{ref}$ and a set of multi-view latent videos $\mathbf{Z}_\text{gen}=\{\mathbf{z}_\text{gen}^{(v)}\}_{v=1}^V$. 
The latent frames are patchified into tokens using a convolutional layer. We remove the text encoder of CogVideoX and retain the diffusion transformer, which then processes these tokens and outputs the predicted noise $\boldsymbol{\epsilon}$~\cite{ho2020denoising}. This transformer jointly attends to all tokens across spatial, frame, and viewpoint dimensions, allowing information to be shared between all pixels of the reference image and all views and frames of the multi-view video. 

We apply spatial and temporal positional encodings to the reference image latent and the video latents following CogVideoX~\cite{yang2024cogvideox}, which uses a sinusoidal positional encoding applied separately to each spatial and temporal dimension.
For the reference image, we use the standard sinusoidal spatial positional encoding, but optimize a learned encoding for the temporal encoding. 
For the video latents, we replicate the same sinusoidal spatial and temporal positional encodings across the viewpoint dimension. 
We adapt the training losses of CogVideoX and fine-tune the model by optimizing all parameters to minimize the L2 difference between the predicted and ground-truth noise added to each latent video frame during the diffusion process. 

\paragraph{Conditioning signals.} 
Following Taubner et al.~\shortcite{taubner2024cap4d}, we condition our model on additional images that provide the head pose, expression, camera pose, motion, and other contextual information for the reference image and the generated multi-view videos (see \cref{fig:method}).
The conditioning signals are produced using a canonical human head model based on FLAME~\cite{li2017learning} and an off-the-shelf head tracker applied to the reference image or to driving videos (i.e., to animate the generated videos)~\cite{taubner2024flowface}. 
A full visualization of the conditioning signals is shown in Figure~S1.

The conditioning signals include: (1) \textit{3D pose maps} $\mathbf{p}_\text{ref}$ and $\mathbf{P}_\text{gen}$, which encode the rasterized canonical 3D coordinates of the head geometry; (2) \textit{expression deformation maps} $\mathbf{e}_\text{ref}$ and $\mathbf{E}_\text{gen}$, which capture the rasterized 3D deformations relative to a neutral expression mesh; (3) \textit{view ray direction and origin maps} $\mathbf{v}_\text{ref}$ and $\mathbf{V}_\text{gen}$, representing the direction and origin of each camera ray in the first camera’s reference frame; and (4) \textit{binary masks} $\mathbf{b}_\text{ref}$ and $\mathbf{B}_\text{gen}$, which indicate whether a frame is a reference or generated frame—$\mathbf{B}_\text{gen}$ is used to condition the model on previously generated video frames.
%
We also include an additional binary mask that identifies regions of the reference image that are padded to accommodate center-cropping (see Supp.\ \cref{supp:sec_mmvdm}).
The conditioning signals for the reference image are given as $\mathbf{c}_\text{ref} = \{\mathbf{p}_\text{ref}, \mathbf{e}_\text{ref}, \mathbf{v}_\text{ref}, \mathbf{b}_\text{ref} \}$, and the signals for the generated images are defined analogously.
We concatenate the reference conditioning signals to the latent reference image $\mathbf{z}_\text{ref}$. 

The conditioning signals for the latent video frames require special handling due to the temporal compression factor.
Following the strategy used in CogVideoX, we retain the conditioning signals for the first input frame and the last frame of each subsequent group of four frames.
To recover high-frequency temporal details lost in this sub-sampling process, we introduce an additional sub-frame motion map, $\mathbf{m}_\text{gen}$, which encodes the 2D screen-space displacement of the 3DMM between the retained frame and each dropped frame within a group (see Supp.\ \cref{supp:sec_mmvdm}).
This allows the model to capture fine-grained motions—such as blinking—that occur over just a few frames.
The compressed conditioning signals are concatenated with their corresponding latent video frames.

\subsection{Multi-Modal Training Curriculum} 
\label{sec:training_curriculum}

Ideally, we would train our proposed multi-view video model on a large-scale multi-view video dataset.
However, no such dataset currently exists.
Furthermore, the number of transformer tokens—and consequently, the computational cost—increases with each additional training view, as each view is associated with a large number of frames.
To address these challenges, we introduce a multi-modal training curriculum for multi-view portrait video generation that enables the synthesis of a large number of frames from multiple viewpoints without training on large-scale multi-view video data.

\begin{table}[t]
    \centering
    \caption{\textbf{Training datasets.} We use datasets spanning different modalities---including monocular videos, dynamic multi-view videos, and static multi-view images---and use them all in our multi-modal training curriculum.}
    \resizebox{\columnwidth}{!}{%
    \begin{tabular}{ l|c c c } 
        \toprule
        dataset & multi-view & video & 360-degree \\ 
        \midrule
        Nersemble~\cite{kirschstein2023nersemble} & \yes & \yes & \no \\
        Ava-256~\cite{ava256} & \yes & \no & \yes \\
        VFHQ~\cite{xie2022vfhq} & \no & \yes & \no \\
        RenderMe-360~\cite{pan2024renderme} & \yes & \yes/\no & \yes  \\
        \bottomrule
    \end{tabular}}
    \label{tab:datasets}
    \Description{This table shows the datasets we used for training the MMVDM and their properties (multi-view, video and 360-degree). }
\end{table}

\paragraph{Datasets.}
Our training curriculum leverages multiple data modalities: monocular, multi-view image, and multi-view video (see \cref{tab:datasets}).
Specifically, VFHQ~\cite{xie2022vfhq} contains 13k clips of forward-facing monocular videos; Nersemble~\cite{kirschstein2023nersemble} contains 8k calibrated forward-facing multi-view videos of subjects performing extreme expressions and speech sequences; RenderMe-360~\cite{pan2024renderme} consists of 5k 360$^\circ$ multi-view image sequences and 500 multi-view video sequences; Ava-256~\cite{ava256} contains 5k 360$^\circ$ multi-view image sets. We use all datasets during all training stages, which are described below.


The multi-view image data provides wide view coverage (e.g., up to 360 degrees), from which the model learns strong multi-view consistency.
The dynamic monocular data from the VFHQ dataset includes diverse motion and identity variations and facilitates temporal consistency.
Finally, the dynamic multi-view data serves as a bridge between the two, helping the model align spatial and temporal cues, despite its limited scale and view coverage.
We use the provided camera calibration information and an off-the-shelf face tracker \cite{taubner2024flowface} to obtain the conditioning signals. 

\paragraph{Generation modes.}
We operate the model in four different modes during training and inference to support flexible multi-view video synthesis conditioned on varying levels of reference information. 
This enables, for example, generating a set of multi-view videos and then using them as references to generate additional videos. In each case, we use the binary masks $\mathbf{B}_\text{gen}$ to indicate which inputs are to be used as references and which are to be generated.

\begin{figure}[t]
\includegraphics[width=\columnwidth]{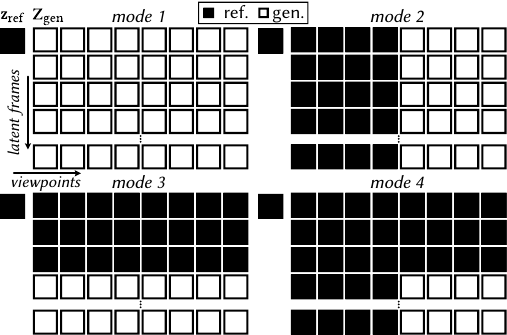}
\caption{Our model supports different modes to generate multiview videos based on the reference image latent $\mathbf{z}_\text{ref}$ and treating different subsets of the latent video frames $\mathbf{Z}_\text{gen}$ as reference inputs.}
\vspace{-1em}
\label{fig:generation-modes}
\end{figure}

The modes are illustrated in \cref{fig:generation-modes} and described as follows.
\begin{itemize}[topsep=0pt,leftmargin=*]
\item \textit{Mode 1} uses only the reference image to generate all views/frames.
\item \textit{Mode 2} conditions on the reference image and half of the viewpoints to generate videos for the remaining viewpoints.
\item \textit{Mode 3} conditions on the reference image and the first three latent frames of all latent video frames to generate additional frames.
\item \textit{Mode 4} combines modes 2 and 3.
\end{itemize}
We train the model on each of these modes, and in \cref{sec:4d-reconstruction} we describe how they are used during inference to generate 4D avatars.

\paragraph{Training curriculum.} 
We leverage our flexible transformer architecture to perform training at varying resolutions across the spatial, viewpoint, and frame dimensions.
Our three-stage training curriculum is outlined in \cref{tab:training_stages}. 
It uses a mixture of datasets and gradually increases the number of output frames, resolution, and generation modes over the training stages.
\begin{itemize}[topsep=0pt,leftmargin=*]
\item Stage 1: we train for 30k iterations at a resolution of $D=256 \times 256$ with different numbers of views $V$, generated frames $F$, and up to one-quarter of the maximum number of total output frames.
\item Stage 2: we train for 40k iterations and increase the resolution to $512 \times 512$ while maintaining the same number of total output frames as stage 1. We also train at $256 \times 256$ resolution with an increased total number of output frames.
\item Stage 3: we train for 5k steps after scaling up to the full number of training views and frames ($V = 4$, $F = 49$).
\end{itemize}
At each stage, we sample all resolutions listed in~\cref{tab:training_stages} with equal probability.
In stage 1, we use mode 1 exclusively.
In stages 2 and 3, if the sampled resolution includes multiple viewpoints, we apply mode 2 with 25\% probability by using half of the viewpoints as reference.
If the resolution includes more than three latent video frames, we apply mode 3 with 25\% probability by using the first three latent frames as reference.
Applying both mode 2 and 3 results in mode 4, and if neither mode is applied, we fall back to mode 1.

This progressive strategy enables efficient training with limited compute while scaling to high spatio-temporal complexity.
At inference time, the model generalizes beyond the number of views and angles seen during training.
Specifically, it synthesizes $V = 8$ synchronized videos with $F = 49$ frames, enabling 360-degree video generation---despite being trained on a very limited number of 360-degree multi-view video data (i.e., 481 sequences~\cite{pan2024renderme}).

\section{4D Avatar Reconstruction}
\label{sec:4d-reconstruction}
We reconstruct a 4D avatar by first using the MMVDM to generate a large set of multi-view videos given the input reference image, and then optimizing a deformable 3D Gaussian splatting (3DGS)-based representation that attaches Gaussian primitives to a 3DMM mesh~\cite{kerbl20233d,qian2024gaussianavatars}.

\paragraph{Sampling.}
We aim to synthesize videos from a large number of viewpoints---up to 48---but generating all of them in a single inference pass is infeasible because (1) the number of viewpoints far exceeds the range the model was trained on, and (2) inference is constrained by available computation and memory. 
To address this, we adopt an iterative generation strategy:
we first produce a small set of key videos from multiple viewpoints, and then progressively expand this set into a coherent multi-view video corpus.
Inference is performed at a spatio-temporal resolution of $(V \times F \times D) = (8 \times 49 \times 512)$.

Concretely, we begin by generating key videos for the 49 output frames (13 latent frames) using generation mode 1.
The remaining target views are grouped into clusters of four neighboring viewpoints, with each cluster generated using mode 2 with the four nearest key videos as reference.
Once all views are synthesized, we apply mode 3 to extend the key videos by generating 40 additional output frames with the last nine output frames from the current key videos as reference.
Finally, the remaining videos are extended using mode 4, which conditions on the four nearest key videos and the last nine frames of the same previously generated videos.
Please see illustrations of this process in Figures S2--S3.

\begin{table}[t]
    \centering
    \caption{\textbf{Training stages.} We gradually increase the viewpoints, frames, resolution, and generation modes at each training stage to improve training efficiency and to leverage different data modalities (monocular video, multi-view videos, and multi-view images).}
    \resizebox{\columnwidth}{!}{%
    \begin{tabular}{ l|c c c } 
        \toprule
        & \multicolumn{3}{c}{training stages}\\
         & stage 1 & stage 2 & stage 3 \\
        \midrule
        \makecell[l]{resolutions \\ ($V \times F \times D$)} & \makecell{1$\times$49$\times$256, 2$\times$29$\times$256, \\ 4$\times$9$\times$256, 12$\times$1$\times$256} & \makecell{1$\times$49$\times$512, 2$\times$29$\times$512, \\4$\times$9$\times$512, 12$\times$1$\times$512, \\ 4$\times$49$\times$256, 8$\times$29$\times$256} & \makecell{1$\times$49$\times$512, 4$\times$49$\times$512, \\ 8$\times$29$\times$512, 16$\times$9$\times$512, \\ 16$\times$1$\times$512}  \\\midrule
        \#iterations & 30k & 40k & 5k  \\ 
        learning rate & 1e-4 & 5e-5 & 2e-5 \\  
        batch size & 64 & 64 & 64 \\\midrule 
        gen.\ mode & 1 & 1--4 & 1--4 \\
        \bottomrule
    \end{tabular}}
    \label{tab:training_stages}
    \vspace{-1em}
     \Description{This table shows the modes and resolutions used at each training stage. }
\end{table}

We generate two kinds of avatars: 120-degree (89 frames per view; 8 key views and 24 other views = 32 views) and 360-degree avatars (89 frames per view; 8 key views and 40 other views = 48 views).
For the 120-degree case, we manually define camera views around the forward-facing head with a max. azimuth angle of 60 degrees and an elevation angle of 30 degrees.
For the 360-degree case, we manually define camera views around the entire head, with a max. and min. elevation angle of +90 and -30.
In each case, we manually select the key views to be positioned uniformly across the view range.

\paragraph{3DGS Fitting.}
We fit a 4D avatar to the generated multi-view videos by adapting the 4D Gaussian splatting representation introduced by Qian et al.~\shortcite{qian2024gaussianavatars} and extended by Taubner et al.~\shortcite{taubner2024cap4d}.
Specifically, we attach Gaussians to the triangles of a FLAME head mesh~\cite{li2017learning}, which is predicted from the reference image using FlowFace~\cite{taubner2024flowface}.
We animate the avatar by deforming the mesh with FLAME blendshapes, using the same expression sequence employed during multi-view video generation.
To model fine-grained effects such as wrinkles, we follow previous work \cite{taubner2024cap4d} and use a U-Net~\cite{ronneberger2015u} to predict frame-dependent, per-Gaussian deformations, represented as a UV map. 

Instead of conditioning this U-Net with an expression-dependent deformation map (as in CAP4D~\cite{taubner2024cap4d}), we condition it on an 8-channel sinusoidal temporal embedding, which allows it to model frame-dependent motion. 

To better capture structures not represented by the FLAME mesh---such as hair, earrings, or glasses---we apply structure-from-motion to keypoints matched across the first frame of each view in the multi-view set. 
We use DISK~\cite{tyszkiewicz2020disk} and LightGlue~\cite{lindenberger2023lightgluelocalfeaturematching} to detect and match keypoints between each key view and the remaining frames.
Keypoints matched in more than three views are triangulated using their screen-space positions, camera poses, and intrinsics to obtain 3D coordinates. 
For each triangulated point, we introduce a Gaussian primitive that is animated with its nearest triangle on the 3DMM mesh.
We optimize the representation to minimize the difference between the rendered views and the generated videos. We regularize the velocity and the relative rotation of Gaussians between each frame (see Sec.\ S2.4).

\section{Implementation Details}

\paragraph{Multi-view classifier-free guidance.}
\sloppy{We apply classifier-free guidance (CFG)~\cite{ho2022classifierfreeguidance} by zeroing out all conditioning signals ($\mathbf{i}_\text{ref}$, $\mathbf{c}_\text{ref}$, and $\mathbf{C}_\text{gen}$) to obtain the unconditional noise prediction $\boldsymbol{\epsilon}_\text{uncond}$.}
The final guided prediction is then computed as:
\begin{equation}
\boldsymbol{\epsilon}_\text{guided} = \boldsymbol{\epsilon}_\text{uncond} + s \cdot (\boldsymbol{\epsilon}_{\text{cond}} - \boldsymbol{\epsilon}_{\text{uncond}}),
\end{equation}
where $s$ is the guidance scale.

\begin{figure*}[ht]
\includegraphics[width=\textwidth]{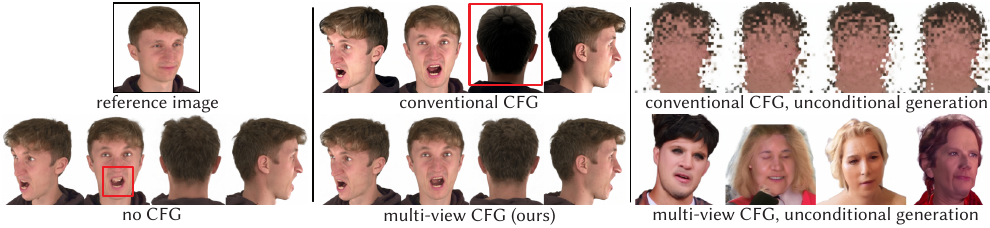}
\vspace{-8mm}
\caption{\textbf{Multi-view classifier-free guidance (CFG).} Conventional CFG (top row, center) causes generations with extreme contrast, due to the poor-quality unconditional prediction (top row, right). Without CFG, artifacts appear (bottom row, left). With our multi-view CFG, the unconditional prediction (bottom row, right) looks plausible, leading to improved generation quality (bottom row, center).}
\label{fig:cfg_figure}
\Description{This figure shows how our multi-view classifier-free guidance method reduces artifacts. Conventional CFG (top row, center) causes generations with extreme contrast, due to the poor-quality unconditional prediction (top row, right). Without CFG, artifacts appear (bottom row, left). With our multi-view CFG, the unconditional prediction (bottom row, right) looks plausible, leading to improved generation quality (bottom row, center). }
\end{figure*}

While standard CFG works well in the single-view setting, it fails when applied naïvely to multiple views. In the unconditional multi-view case, all tokens receive identical spatio-temporal embeddings and noisy latent patches, without any view-specific information. This ambiguity causes the diffusion transformer to confuse token relationships across views, resulting in poor unconditional predictions.
To address this, we generate view-specific unconditional predictions by running independent forward passes for each predicted view:
\begin{equation}
\boldsymbol{\epsilon}_\text{uncond} = \{ \boldsymbol{\epsilon}_{\text{uncond},v} \}_{v=1}^V,
\end{equation}
where $\boldsymbol{\epsilon}_\text{uncond,v}$ is the unconditional prediction for view $v$. We show in our ablations that this improves quantitative performance, with further qualitative examples provided in \cref{fig:cfg_figure}. 
Following previous work \cite{kant2025pippo}, we linearly increase the guidance scale with the angular distance to the closest reference view, from 1.1 (0 degree deviation) to 1.6 (180 degree deviation).

\paragraph{Attention biasing.}
The total number of tokens processed by the attention layers in the diffusion transformer scales with the number of views, spatial resolution, and video frames.
Due to the combined compression from the auto-encoder and token patchification, the input is downsampled by a factor of 16 in both height and width, and approximately 4 in time, resulting in a total token count of $([1+ V \times (1+\tilde{F})]\times \frac{H}{16} \times \frac{W}{16})$.
Prior work~\cite{jin2023trainingfree} shows that the entropy of attention grows logarithmically with the number of tokens, and proposes to compensate for this increase by scaling the attention weights during inference as:
$\sqrt{\frac{1}{d}} \sqrt{ \frac{\log N_i}{\log N_t} }$,
where $N_i$ is the number of tokens at inference, $N_t$ is the number of tokens used during training, and $d$ is the token dimensionality.

We adopt this approach by applying attention scaling during both training and inference.
Specifically, we set $N_i$ to the token count for each individual batch and  $N_t$ to the default configuration of the CogVideoX model ($13\times32\times32$).
For further discussion on attention biasing and its effects, we refer the reader to Kant et al.~\shortcite{kant2025pippo}.

\paragraph{Training \& Inference.}
The total training time is 14 days on 8$\times$H100 GPUs. We drop out all conditioning signals with a probability of 10\%.
We use DDIM sampling~\cite{song2021denoising} during inference with 100 sampling steps for the key views and 50 steps for all other views. 
Generating 32 views and 89 video frames (120-degree avatar) takes around 6.5 hours on a single RTX 6000 Ada GPU.  Generating 48 views and 89 video frames takes around 10.5 hours. 4D reconstruction takes 1.5 hours. After fitting, rendering the sequence with a resolution of 512$\times$512 is real-time. 

\section{Experiments}


\paragraph{Baselines.}
We evaluate our method against several baselines for 3D representation-based single-view avatar reconstruction: GAGAvatar~\cite{chu2024gagavatar}, Portrait4D-v2~\cite{deng2024portrait4dv2}, and VOODOO3D~\cite{voodoo3d} and CAP4D~\cite{taubner2024cap4d}. 
We also include CAP4D’s morphable multi-view diffusion model (MMDM), which synthesizes images across viewpoints and expressions.
Additionally, we combine the portrait video generation model Follow-Your-Emoji (FYE)~\cite{ma2024followyouremoji} with the static multi-view generation model PanoHead~\cite{an2023panohead} to generate 360-degree 4D avatars.
We refer to Supp.\ \cref{supp:sec_self_reenactment} for additional baseline details.

\paragraph{Metrics.}
We compare each method using several key metrics: photometric accuracy (PSNR, SSIM, LPIPS), temporal consistency (JOD)~\cite{mantiuk2021jod}, and identity preservation, measured using the cosine similarity of identity embeddings (CSIM)~\cite{deng2022arcface}.
To assess 3D consistency, we compute the reprojection error (RE@LG) of DISK~\cite{tyszkiewicz2020disk} keypoints using known camera poses and predicted correspondences from LightGlue~\cite{lindenberger2023lightgluelocalfeaturematching}, following Kant et al.~\shortcite{kant2025pippo}. 

\newcommand{\best}[1]{\textbf{#1}}
\newcommand{\second}[1]{\underline{#1}}
\newcommand{\third}[1]{#1}

\begin{table}[t]
    \centering
    \caption{\textbf{Self-reenactment results on the Nersemble dataset.} When generating multi-view video, MVP4D outperforms previous methods on the JOD metric, while performing competitively on photometric accuracy (PSNR), which indicates we achieve improvements in temporal consistency.}
    \vspace{-1em}
    \resizebox{\columnwidth}{!}{%
    \begin{tabular}{ l|c c c c c c c} 
        \toprule
        Method & PSNR$\uparrow$ & SSIM$\uparrow$ & LPIPS$\downarrow$ & JOD$\uparrow$ & RE@LG$\downarrow$ & CSIM$\uparrow$ \\ 
        \midrule
        GAGAvatar     & 20.01 & 0.721 & 0.353 & 4.73 & 2.25  & 0.656  \\ 
        Portrait4D-v2 & 17.05 & 0.662 & 0.392 & 3.67 & 3.75  & 0.659  \\
        VOODOO3D      & 20.09 & 0.692 & 0.349 & 4.86 & 1.84  & 0.517  \\ 
        CAP4D (MMDM)  & 22.17 & 0.768 & \best{0.280} & 5.30 & 1.28  & \best{0.793} \\ 
        CAP4D (120)   & 21.56 & 0.778 & 0.302 & 5.49 & \best{0.634} & \second{0.755}  \\ 
        \midrule
        MVP4D (MMVDM) & \best{23.39} & 0.765 & 0.294 & \best{5.88} & 1.03  & 0.707 \\ 
        MVP4D (360)   & 22.76 & 0.774 & 0.308 & 5.57 & \second{0.643} & 0.715  \\ 
        MVP4D (120)   & \second{23.16} & \best{0.790} & \second{0.301} & \second{5.62} & 0.656 & 0.708 \\ 
        \bottomrule
    \end{tabular}}
    \label{tab:self_reenactment_results_forward}
    \Description{Quantitative results for the self-reenactment evaluation on the Nersemble dataset. When generating multi-view video, MVP4D outperforms previous methods on the JOD metric, while performing competitively on photometric accuracy, which indicates we achieve improvements in temporal consistency.}
\end{table}

\begin{table}[t]
    \centering
    \caption{\textbf{Self-reenactment results on the RenderMe-360 dataset.} \method outperforms other baselines for 360-degree self-reenactment.}
    \vspace{-1em}
    \begin{tabular}{ l|c c c c } 
        \toprule
        Method & PSNR$\uparrow$ & SSIM$\uparrow$ & LPIPS$\downarrow$ & JOD$\uparrow$ \\
        \midrule
        F-Y-E + PanoHead & 15.31 & 0.53 & 0.451 & 3.14  \\
        CAP4D (MMDM) & 13.89 & 0.58 & 0.387 & 3.19 \\ 
        \midrule
        MVP4D (MMVDM) & 18.77 & 0.68 & \best{0.321} & 4.37  \\  
        MVP4D (360) & \best{18.78} & \best{0.69} & 0.323 & \best{4.45} \\ 
        \bottomrule
    \end{tabular}
    \label{tab:self_reenactment_results_360}
    \vspace{-1em}
    \Description{Quantitative results for the self-reenactment evaluation on the RenderMe-360 dataset. MVP4D outperforms other baselines for 360-degree self-reenactment.}
\end{table}

\subsection{Self-reenactment}
We benchmark self-reenactment on ten forward-facing multi-view sequences from the Nersemble dataset~\cite{kirschstein2023nersemble} and eight 360-degree sequences from the RenderMe-360 dataset~\cite{pan2024renderme}.
For Nersemble, we select ten sequences featuring extreme facial expressions and speech, using the first 70 frames (captured at 24 FPS) from 8 of the 16 available viewpoints.
For RenderMe-360, we select eight sequences, sampling 70 frames (captured at 15 FPS) from 8 of the 60 viewpoints in each.
In both cases, the reference image is taken from a different frame and viewpoint than those used for evaluation, and we assess how accurately each method reenacts the target subject’s appearance.

\begin{figure*}[ht]
\includegraphics[width=\textwidth]{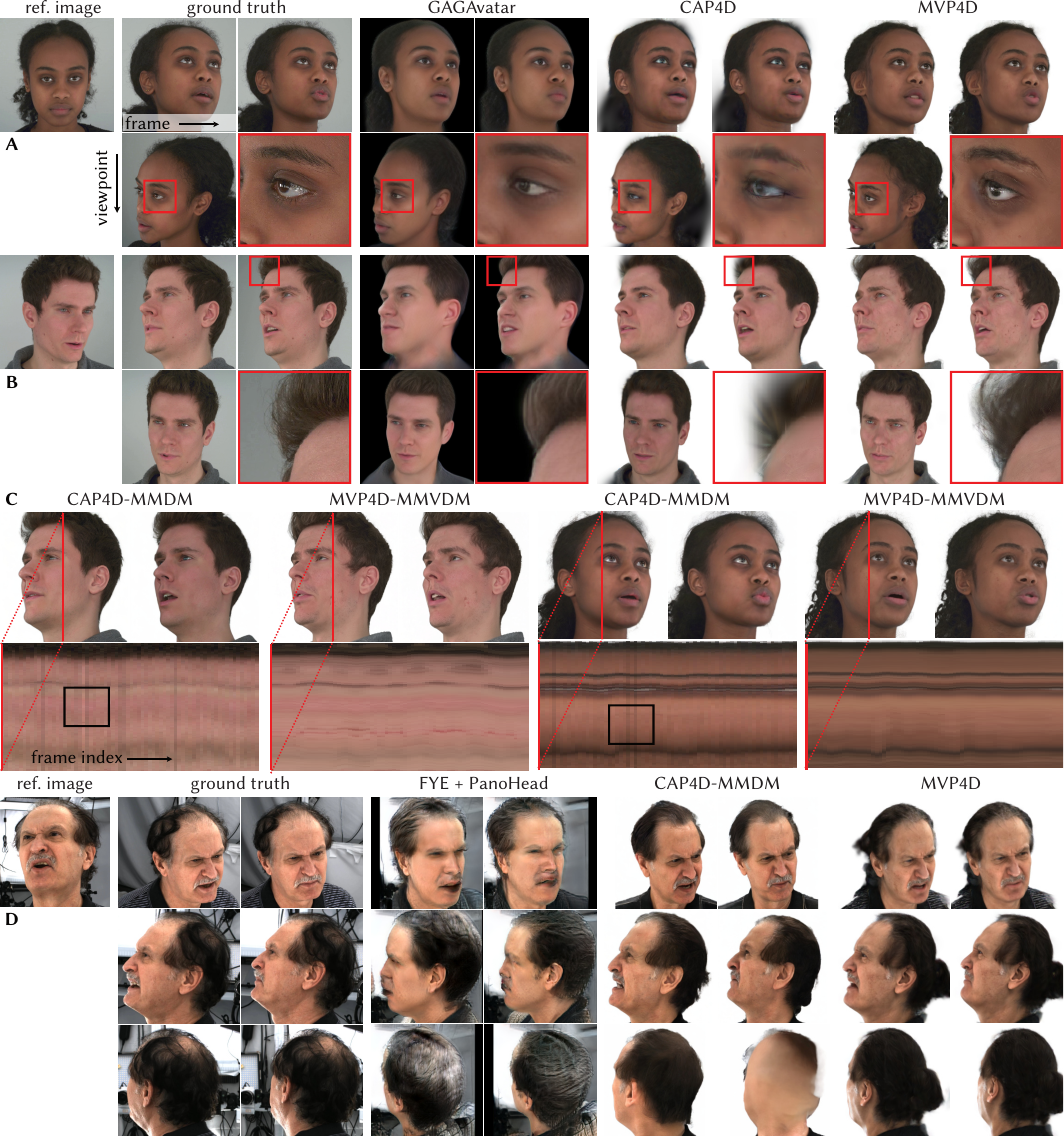}
\vspace{-2em}
\caption{\textbf{Self-reenactment results on the Nersemble and RenderMe-360 datasets.} 
Each method generates multi-view videos from a single reference image (left). 
(\textbf{A}, \textbf{B}) We compare the outputs of \method and baselines that support real-time rendering of the 4D avatar. \method recovers fine details and structures of the face and hair that are not captured by other techniques (see insets).
(\textbf{C}) CAP4D-MMDM and MVP4D-MMVDM generate views from a single reference image using a diffusion model. However, the CAP4D-MMDM output flickers because it relies on an image diffusion model. 
The flickering appears as high-frequency artifacts (black boxes) in the space-time visualization shown for a column of pixels in the generated videos (red lines).
(\textbf{D}) Finally, we show 360-degree generation on the RenderMe-360 dataset. \method outperforms baselines for this task, which fail or have noticeable artifacts, especially when generating the back of the head.}
\label{fig:self_reenactment_qualitative}
\Description{Qualitative results from the self-reenactment evaluation on the Nersemble and RenderMe-360 datasets. Generated images for MVP4D and multiple baselines are shown next to the corresponding reference and ground truth images. }
\end{figure*}

\paragraph{Forward facing avatar evaluation.}
We present results in Tables~\ref{tab:self_reenactment_results_forward}--\ref{tab:self_reenactment_results_360} and Figure~\ref{fig:self_reenactment_qualitative}.
\method outperforms the baselines in temporal consistency, as measured by JOD~\cite{mantiuk2021jod}, and performs competitively in terms of photometric accuracy (PSNR, SSIM) (see \cref{tab:self_reenactment_results_forward}).
While CAP4D-MMDM achieves better perceptual similarity (LPIPS), this metric is more tolerant of geometric inconsistencies between the generated and ground truth images.
The stronger temporal consistency of our method suggests that, although MMDM may produce higher-quality individual frames, our approach yields more temporally coherent results.
This advantage likely stems from our model generating all views simultaneously (8~$\times$~49 frames), in contrast to CAP4D-MMDM, which synthesizes only eight views at a time.
Qualitatively, \method produces videos with improved temporal stability (less flickering) compared to CAP4D-MMDM, and with higher visual fidelity than GAGAvatar, as shown in \cref{fig:self_reenactment_qualitative}. We show additional metrics and baselines in Supp.\ \cref{supp:sec_self_reenactment}.

\paragraph{360-degree avatar evaluation.}
CAP4D-MMDM is not trained on 360-degree data and fails to generate views behind the head. 
Both FYE+PanoHead and CAP4D struggle to generate consistent geometry in areas that are not visible from the reference view, since they do not generate all views at once. This is noticeable in qualitative results (see the project webpage), e.g., in the form of a flickering hairstyle. FYE+PanoHead further suffers from artifacts because of the two-step generation (first monocular video, then multi-view). 

\subsection{Cross-reenactment}

We evaluate cross-reenactment using 10 reference images from the FFHQ dataset~\cite{karras2019ffhq}, each paired with a driving video from the Nersemble dataset~\cite{kirschstein2023nersemble}, comprising 8 sequences with extreme expressions and 2 with speech.
For each reference image, we reconstruct a 4D avatar and animate it using the corresponding driving video.
We render each reenacted sequence with a camera following an elliptical trajectory around the head, sweeping up to 60° in azimuth and 20° in elevation.

To assess perceptual quality, we conducted a user study comparing our method (MVP4D) to baselines.
We collected 6,750 responses from 45 participants, each shown the reference image, driving video, and side-by-side results from MVP4D and a baseline.
Participants selected their preference based on five criteria: visual detail (VD), expression transfer accuracy (ET), 3D structure fidelity (3DS), motion quality (MQ), and overall preference (Overall).
As shown in \cref{tab:cross_reenactment_quanitative_main}, MVP4D was preferred overall in 86\% of cases compared to the strongest baseline, CAP4D. 
We provide additional details about the user study in Supp.\ \cref{supp:sec_cross_reenactment}.

\subsection{Ablations and Extensions}

\paragraph{Ablation study.}
All ablation studies are conducted on the Nersemble self-reenactment evaluation set to assess the impact of multi-view CFG and the number of jointly generated views.
As shown in Table~\ref{tab:ablations}, omitting CFG entirely (i.e., unconditional generation) results in a substantial drop in generation quality.
Applying standard CFG without our proposed multi-view strategy also yields subpar performance.
We also analyze the effect of the number of jointly generated views $V$ during diffusion sampling.
Although the model was trained with $V=4$ views, we observe improved performance when sampling with $V=8$ views---even without explicit training in this configuration.
We find that jointly generating a larger set of views ($V$=$8$) improves the 3D consistency and quality of the resulting multi-view video corpus.
We validate our multi-modal training curriculum with additional ablation studies in Supp. \cref{supp:sec_multi_training}, and show how generation longer videos degrades performance in Supp. \cref{supp:sec_long_video}.

\begin{figure*}[ht]
\includegraphics[width=\textwidth]{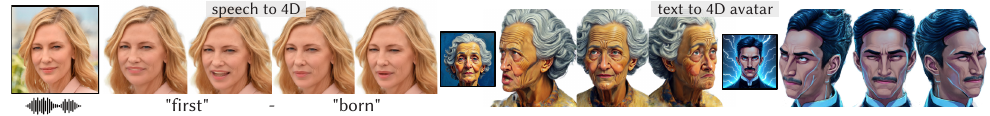}
\vspace{-6mm}
\caption{\textbf{More results.} Using off-the-shelf speech-driven video animation methods, MVP4D can generate 4D portrait avatars from audio input and a reference image. MVP4D can also generate 4D avatars from generated images (please see videos in the project page). }
\label{fig:bells_and_whistles}
\Description{This figure shows how MVP4D can also generated 4D avatars from generated images. }
\end{figure*}

\paragraph{Speech-Driven 4D Avatar Generation.}
Given a reference image and input speech audio, we use Hallo3~\cite{cui2024hallo3} to generate an animated portrait video.
We then apply a face tracker to this video to extract 3DMM-based conditioning signals.
Our model is conditioned on the reference image and the extracted 3DMM from the driving video (see Figure~\ref{fig:bells_and_whistles}).

\paragraph{Text-to-4D avatar.}
We generate 4D avatars from text prompts by first generating an image with a commercial image generation model~\cite{blackforestlabs2024flux1} and then applying \method.

\begin{table}[t]
    \setlength{\tabcolsep}{3pt}
    \captionof{table}{\textbf{Cross-reenactment results.} We evaluate human preference based on visual detail (VD), expression transfer (ET), 3D structure (3DS), motion quality (MQ), and overall preference (Overall). The table reports the percentage of users (45 participants) who preferred MVP4D over the corresponding baseline in side-by-side comparisons.}
    \vspace{-1em}
    \begin{center}
    \begin{tabular}{ l|c c c c c } 
        \toprule
        & \multicolumn{5}{c}{human preference (ours vs. method)} \\
        Method  & VD & ET & 3DS & MQ & Overall  \\
        \midrule
        Portrait4D-v2 & 94\% & 94\% & 92\% & 95\% & 93\% \\
        GAGAvatar     & 97\% & 87\% & 88\% & 87\% & 88\% \\
        CAP4D         & 91\% & 80\% & 85\% & 81\% & 82\% \\
        \bottomrule
    \end{tabular}
    \end{center}
    \label{tab:cross_reenactment_quanitative_main}
    \Description{Quantitative result for the cross-reenactment evaluation. We evaluate human preference based on visual detail, expression transfer, 3D structure, motion quality, and overall preference. The table reports the percentage of 45 participants who preferred MVP4D over the corresponding baseline in side-by-side comparisons. The results favor MVP4D, with a 82 percent preference over the best baseline.}
    \vspace{-1em}
\end{table}

\begin{table}[t]
    \centering
    \caption{\textbf{Ablation studies.} 
   We assess performance across CFG modes and number of generated views $V$. Omitting CFG or using conventional CFG reduces performance. Although the model is trained with $V=4$, the best quality and 3D consistency is achieved with $V=8$ and multi-view CFG.}
   \vspace{-1em}
    \resizebox{\columnwidth}{!}{%
    \begin{tabular}{lc|c c c c } 
        \toprule
        CFG mode & $V$ & PSNR$\uparrow$ & LPIPS$\downarrow$ & JOD$\uparrow$ & RE@LG$\downarrow$ \\
        \midrule
        no CFG & 8 & 23.34 & 0.297 & \best{5.58} & 1.24 \\  
        conventional CFG & 8 & 22.27 & 0.308 & 5.44 & 1.04  \\ 
        multi-view CFG & 4 & 22.55 & \best{0.291} & 5.49 & 1.73 \\
        \midrule
        multi-view CFG & 8 & \best{23.39} & 0.294 & \best{5.58} & \best{1.03}  \\ 
        \bottomrule
    \end{tabular}}
    \label{tab:ablations}
    \Description{We assess performance across CFG modes and number of generated views V. Omitting CFG or using conventional CFG reduces performance. Although the model is trained with V equals 4, the best quality and 3D consistency is achieved with V equals 8 and multi-view CFG.}
    \vspace{-1em}
\end{table}

\section{Discussion}
\begin{figure*}[ht]
\includegraphics[width=\textwidth]{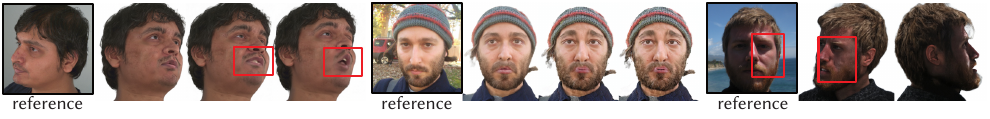}
\vspace{-8mm}
\caption{\textbf{Limitations.} Artifacts in high frequency spatio-temporal details (appearing and disappearing lips) are likely due to the low spatio-temporal resolution in the latent space (left). When predicting very long sequences, quality deteriorates over time (due to conditioning on previous, non-perfect frames; middle). The model may also fail under extreme lighting (right), as our training procedure does not incorporate such data.}
\label{fig:limitations}
\Description{This figure shows limitations of MVP4D. Artifacts in high frequency spatio-temporal details (appearing and disappearing lips) are likely due to the low spatio-temporal resolution in the latent space (left). When predicting very long sequences, quality deteriorates over time (due to conditioning on previous, non-perfect frames; middle). The model may also fail under extreme lighting (right), as our training procedure does not incorporate such data. }
\end{figure*}

Our work highlights the promise of emerging video diffusion transformers for generating realistic and temporally coherent 4D avatars.
While our approach outperforms previous techniques in terms of visual fidelity and temporal consistency, we also illustrate some limitations of \method in \cref{fig:limitations}.
For example, due to the auto-encoder's high spatiotemporal compression, our method struggles to capture fine-grained temporal details, such as rapidly appearing and disappearing teeth.
When generating long sequences using the first-frame conditioning strategy (generation mode 3), quality degrades with each iteration, limiting the practical sequence length.
Additionally, the method cannot accurately model extreme lighting conditions, as the training data lacks significant lighting variation in the multi-view setting.

Hence, there are many exciting avenues for future research.
For example, integrating physically based lighting and reflectance models could enable more accurate control over avatar illumination.
Additionally, building on recent methods that directly generate 3D representations in a feed-forward manner~\cite{szymanowicz2025bolt3dgenerating3dscenes} may enable more efficient 4D avatar generation.


\begin{acks}
    
DBL acknowledges support from LG Electronics, the Natural Sciences and Engineering Research Council of Canada (NSERC) under the RGPIN, RTI, and Alliance programs, the Canada Foundation for Innovation, and the Ontario Research Fund. The authors also acknowledge computing support provided by the Vector Institute.

\end{acks}

\bibliographystyle{ACM-Reference-Format}
\bibliography{ref}

\clearpage

\begin{figure*}[ht]
\includegraphics[width=\textwidth]{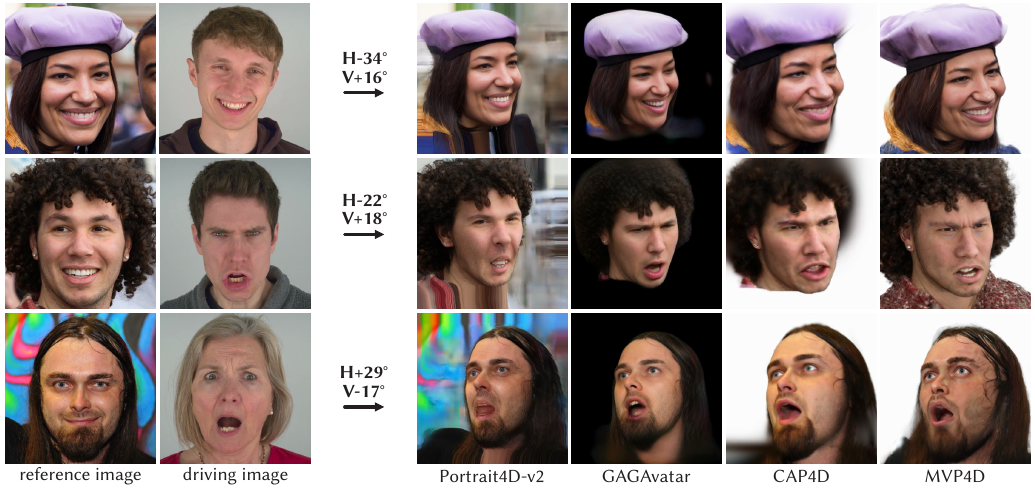}
\vspace{-8mm}
\caption{\textbf{Cross-reenactment results.} Our method reconstructs challenging geometry, such as hair, significantly better than previous methods, while also modeling detailed dynamic effects such as wrinkles.}
\label{fig:cross_reenactment_qualitative}
\Description{Qualitative results from the cross-reenactment evaluation. Generated images for MVP4D and multiple baselines are shown next to the corresponding reference and driving images. }
\end{figure*}

\let\balance\relax
\let\endbalance\relax
\FloatBarrier

\clearpage
\appendix

\renewcommand{\thesection}{S\arabic{section}}
\renewcommand{\thefigure}{S\arabic{figure}}
\renewcommand{\thetable}{S\arabic{table}}
\renewcommand{\theequation}{S\arabic{equation}}

\twocolumn[
  \section*{\Huge Multi-View Portrait Video Diffusion for Animatable 4D Avatars}
  \vspace{0.2em}
  \section*{\Huge Supplemental Material}
  \vspace{1em}
]

\noindent
This document includes supplementary implementation details and results. 
We provide implementation details related to the morphable multi-view video diffusion model (MMVDM), 4D reconstruction, datasets, and evaluation procedures.
\textbf{Please also refer to our project page for code, video results, and comparisons to baselines \href{https://felixtaubner.github.io/mvp4d/}{\textcolor{blue}{\texttt{https://felixtaubner.github.io/mvp4d/}}}}.
\bigskip


\section{Ethics Statement}
We acknowledge that although our video model can generate realistic human avatars, there is still a gap in realism. Still, we condemn any misuse of generated avatars to manipulate or harm individuals, communities, or entities. Moreover, we emphasize the importance of transparency and responsible use as the technology advances as part of future work. 

\section{Supplemental Implementation Details}

\subsection{Morphable Multi-view Video Diffusion Model}
\label{supp:sec_mmvdm}

\paragraph{Architecture details}

\begin{figure*}[ht!]
\includegraphics[width=0.94\textwidth]{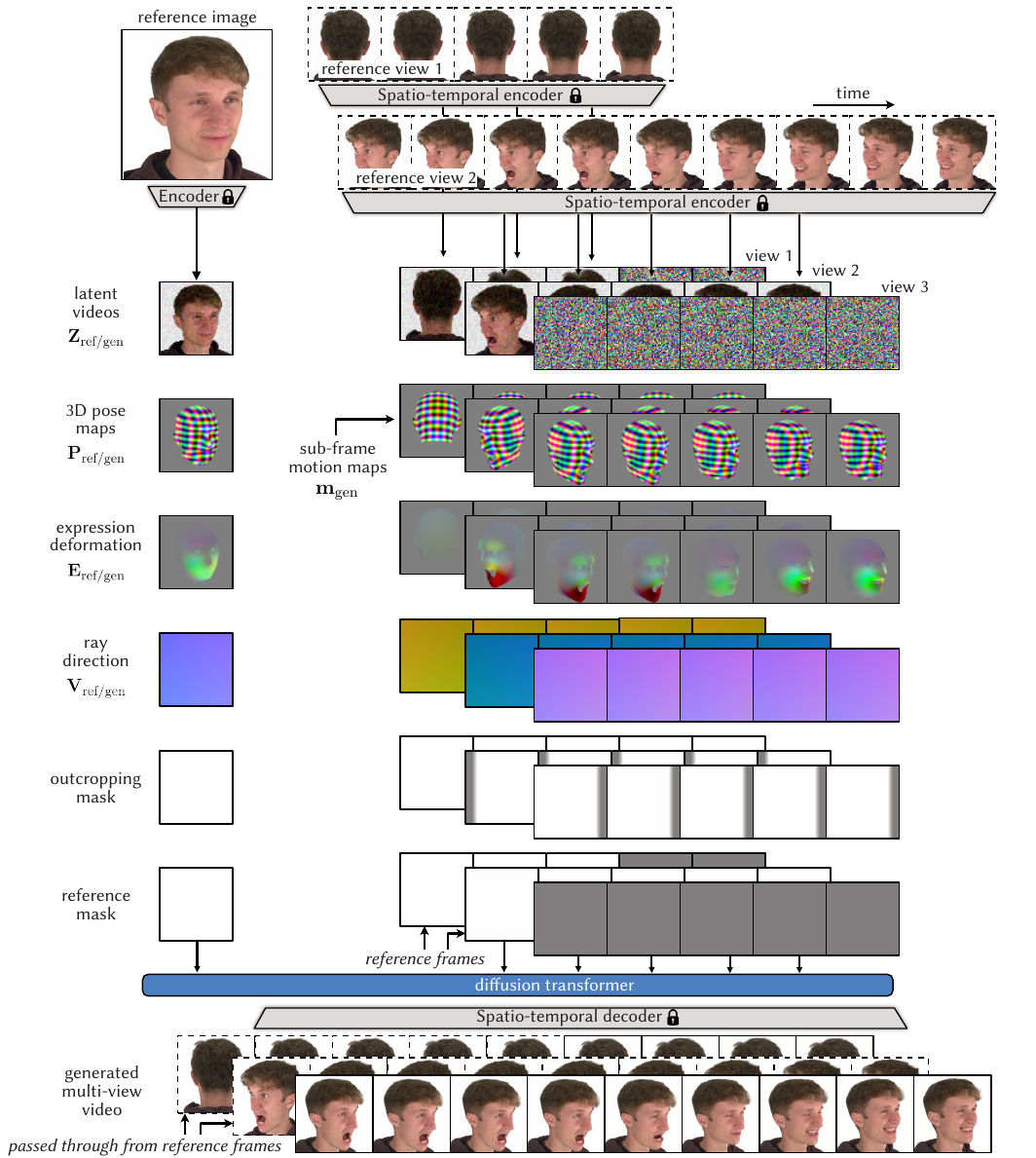}
\vspace{-1em}
\caption{\textbf{Full visualization of the conditioning signals and the MMVDM architecture.} The model synthesizes multi-view videos from a single input reference image. We encode the input reference image into a latent frame, and we provide the model with a set of conditioning signals, noisy latents, and, optionally, an additional set of generated video frames to facilitate auto-regressive generation of videos. 
The model is conditioned on signals including: (1) \textit{3D pose maps} $\mathbf{p}_\text{ref}$ and $\mathbf{P}_\text{gen}$, which encode the rasterized canonical 3D coordinates of the head geometry; (2) \textit{expression deformation maps} $\mathbf{e}_\text{ref}$ and $\mathbf{E}_\text{gen}$, which capture the rasterized 3D deformations relative to a neutral expression mesh; (3) \textit{view ray direction and origin maps} $\mathbf{v}_\text{ref}$ and $\mathbf{V}_\text{gen}$, representing the direction and origin of each camera ray in the first camera’s reference frame; and (4) \textit{binary masks} $\mathbf{b}_\text{ref}$ and $\mathbf{B}_\text{gen}$, which indicate whether a frame is a reference or generated frame—$\mathbf{B}_\text{gen}$ is used to condition the model on previously generated video frames. Following Taubner et al.~\shortcite{taubner2024cap4d}, we also include an outcropping mask, which denotes the valid region of a padded video frame. The conditioning signals, noisy latents, and reference latents are passed into the diffusion transformer, which is used to recover denoised latent multi-view videos. The result is passed through a decoder to recover the output multi-view video sequence.}
\label{supp:mmvdm_conditioning}
\end{figure*}

\sloppy{We show a detailed illustration of the MMVDM architecture in \cref{supp:mmvdm_conditioning}. The reference image and the reference views are projected into the low-resolution latent space using the pretrained CogVideoX auto-encoder \cite{yang2024cogvideox}, with a spatial downsampling ratio of 8 and a temporal downsampling ratio of 4.} Single images (such as the reference image) are processed individually to produce one latent frame. Then, all conditioning maps are concatenated to the latent frames before patchification into tokens with a latent spatial patch size of $2 \times 2$. The tokens from the reference latent ($1\times \frac{H}{16} \times \frac{W}{16}$ tokens), and each view (each with $(1 + \tilde{F}) \times \frac{H}{16} \times \frac{W}{16}$ tokens) are then flattened and concatenated, leading to a total number or $([1+ V \times (1+\tilde{F})]\times \frac{H}{16} \times \frac{W}{16})$ in the transformer. The 30-layer diffusion transformer processes all tokens simultaneously, effectively sharing information across all frames, pixels, and views.

\paragraph{Conditioning details.}

We use a background-matting model~\cite{lin2021robustvideomatting} to mask the reference image to a white background.
We adapt our conditioning signals from CAP4D \cite{taubner2024cap4d}. For a detailed explanation of the 3D pose maps $\mathbf{p}_\text{ref}$, expression deformation maps $\mathbf{e}_\text{ref}$, view rays $\mathbf{v}_\text{ref}$, binary reference masks $\mathbf{b}_\text{ref}$ and outcropping masks, we refer to their paper. We modify their conditioning maps in a few ways:

\paragraph{Increased face bounding box.}
We increase the crop region around the face to 25\% larger (from 20\%) than a tight-fitting bounding box around the projected head vertices. We also apply a temporal Gaussian smoothing filter to the bounding box position to avoid excessive jittering. 

\paragraph{Camera origin conditioning.}
We append a three-channel map to the view ray direction map $\mathbf{v}_\text{ref}$, indicating the camera origin of each view, relative to the reference view. 

\paragraph{Sub-frame motion maps.}
To account for the compression factor along the frame dimension, we drop the conditioning signals for all frames except the first input frame and the last frame of each subsequent group of four frames.
To recover high-frequency temporal details (e.g., blinking) lost in this sub-sampling process, we introduce an additional sub-frame motion map, $\mathbf{m}_\text{gen}$, which encodes the 2D screen-space displacement of the 3DMM between the retained frame and each dropped frame within a group. This 2D screen-space displacement is rasterized into an image in the same way as $\mathbf{e}_\text{ref}$. This results in three two-channel motion maps for each group of four frames (none for the reference frame and first frame of the video), which is also concatenated to the conditioning signals and latent. 

\subsection{Training}
As mentioned in \cref{sec:training_curriculum}, we use a collection of four datasets for our training: 
VFHQ~\cite{xie2022vfhq}, Nersemble~\cite{kirschstein2023nersemble}, RenderMe-360~\cite{pan2024renderme} and Ava-256~\cite{ava256}. 
For all datasets, we use FlowFace~\cite{taubner2024flowface} and a gaze estimation model~\cite{abdelrahman2022l2cs} to obtain FLAME~\cite{li2017learning} annotations. For Nersemble, RenderMe-360 and Ava-256, we use the available camera calibrations during tracking. FlowFace cannot detect back-facing heads; however, with the provided camera calibration parameters in the dataset, the 3DMM pose relative to back-facing cameras can still be obtained. 
We remove the background using RobustVideoMatting~\cite{lin2021robustvideomatting}. 
Some sequences of the VFHQ dataset contain video transitions. We detect these transitions by measuring the average keypoint acceleration (with the FlowFace keypoint detector) above a certain threshold, and discard them. The VFHQ dataset also occasionally contains frames where hands or objects occlude the face, however we do not remove those.

To train our model, we randomly select $F$ frames and $V$ views from each dataset video sequence, depending on the resolutions of the current training stage, and on the available number of views $V_\text{max}$ and frames $F_\text{max}$ of the particular sequence. 
For example, AVA-256 sequences and most sequences of RenderMe-360 are not temporally coherent but contain many views ($F_\text{max}=1$ and $V_\text{max}>16$). The Nersemble dataset and some sequences of RenderMe-360 are multi-view and temporally coherent ($F_\text{max}=49$ and $V_\text{max} \geq 16$). VFHQ contains only monocular videos ($F_\text{max}=49$ and $V_\text{max} = 1$). 
For the reference image, we randomly select a camera view from all views where a face is detected and the head is facing the camera; from this view, we randomly select a frame from the remaining frame indices that are not contained in the multi-view evaluation sequences. 
\clearpage


\subsection{Sampling}

\begin{figure*}[ht!]
\includegraphics[width=\textwidth]{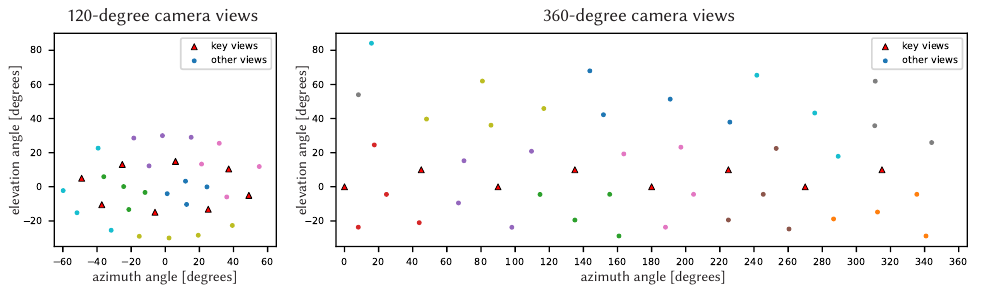}
\caption{\textbf{The sampled camera view angles} for 120-degree avatars (left) and 360-degree avatars (right). Azimuth angle (x-axis) and elevation angle (y-axis) are defined relative to the head direction. The key views are denoted as triangles. Other views are denoted as points, while their color illustrates their cluster assignments. Each cluster is generated conditioned on the 4 key views that are closest to the corresponding cluster center. }
\label{fig:supp_sampling_camera_distr}
\end{figure*}

\begin{figure*}[ht]
\includegraphics[width=\textwidth]{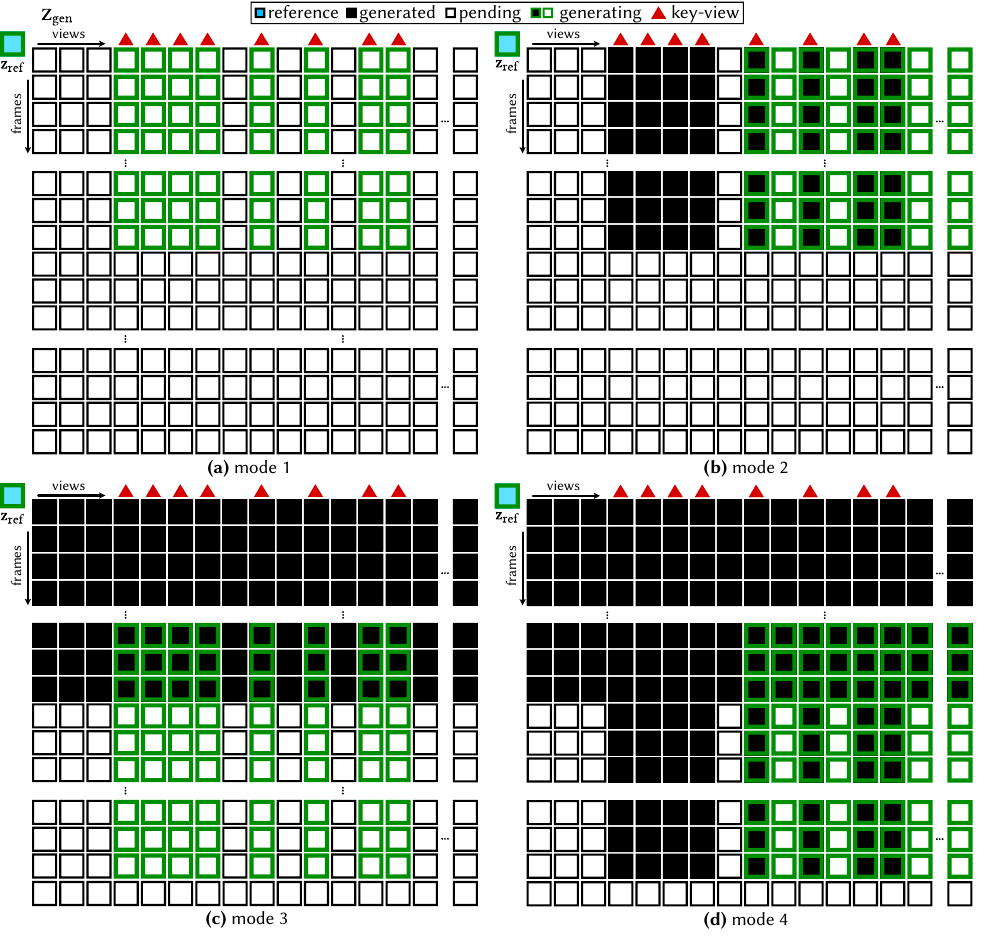}
\vspace{-8mm}
\caption{\textbf{Illustration of our step-by-step multi-view video sampling strategy.} 
 (a) We begin by generating key videos for the first 49 output frames (13 latent frames) using generation mode 1.
 (b) The remaining target views are grouped into clusters of four neighboring viewpoints, with each cluster generated using mode 2 with the four nearest key videos as reference.
 (c) Once all views are synthesized, we apply mode 3 to extend the key videos by generating 40 additional output frames with the last nine output frames from the current key videos as reference.
 (d) Finally, the remaining videos are extended using mode 4, which conditions on the four nearest key videos and the last nine frames of the same previously generated videos.
}
\label{fig:supp_sampling_procedure}
\vspace{-3mm}
\end{figure*}

To create the 4D avatar, we generate a large collection of camera views around the head using the diffusion model. These views are manually defined to be at certain angles (azimuth and elevation) relative to the forward-facing direction of the head. See \cref{fig:supp_sampling_camera_distr} for a detailed view of how we select the sampled camera view angles (key views and other views) for the 120-degree and 360-degree avatar generation case. See \cref{fig:supp_sampling_procedure} for an illustration of how these views and all timesteps are generated in a step-by-step process.

\subsection{4D Reconstruction Details}
\label{supp:avatar_reconstruction_details}

To reconstruct the 4D sequence, we use a 4D representation largely based on GaussianAvatars~\cite{qian2024gaussianavatars}, which is extended by CAP4D~\cite{taubner2024cap4d}. These method attach Gaussian primitives to the 3DMM mesh, which is driven by the reference head shape and the input animation sequence. 
We animate the avatar by deforming the mesh with FLAME blendshapes, using the same expression sequence employed during multi-view video generation. 
We follow CAP4D's implementation, with a few minor modifications to better model detailed temporal dynamics, as follows.

\paragraph{Temporal positional embedding and frame-dependent deformation.}
CAP4D remeshes the FLAME template mesh into UV-pixel-aligned vertices and faces, and deforms them using a U-Net~\cite{ronneberger2015u} to predict expression-dependent, per-Gaussian deformations, in UV space. Since we are overfitting the representation to the generated sequence, we remove the expression-map conditioning and replace it with an 8-channel sinusoidal positional encoding. We keep the spatial positional encoding of CAP4D. The U-Net thus outputs the \textbf{time-dependent} UV-deformation map for each frame index. 

\paragraph{Velocity regularization.}
To prevent excessive motion of the Gaussians introduced by the U-Net, we apply a regularization of the relative rotation and position of each Gaussian between subsequent timesteps during training.
We remove the relative deformation and rotation losses $\mathcal{L}_\text{deform}$ and $\mathcal{L}_\text{rot}$ between deformed and canonical poses of each Gaussian and replace them with similar, velocity-based losses to penalize excessive motion.
The relative rotation loss is computed as the square of the magnitude of the axis-angle rotation $R$ between the $i$th Gaussian at the current frame $f$ and the next frame $f+1$:
\begin{equation}
    \mathcal{L}_{\text{v,rot}} = \lambda_\text{v,rot} \frac{1}{N} \sum_{i=1}^N \left\| \log\left( R_{i,f}^\top R_{i,f+1} \right) \right\|_2^2.
\end{equation}
The relative deformation loss is computed as the square of the magnitude of the difference (velocity) between the U-Net induced deformation $d$ of each Gaussian $i$ at the current timestep $t$ and the next timestep $t+1$:
\begin{equation}
    \mathcal{L}_{\text{v,deform}} = \lambda_\text{v,deform} \frac{1}{N} \sum_{i=1}^N  \left\| d_{i,t+1} - d_{i,t} \right\|_2^2
\end{equation}
We follow the training strategy of CAP4D in all other aspects, keeping the loss functions from GaussianAvatars and CAP4D, including LPIPS loss \cite{zhang2018lpips} $\mathcal{L}_\text{LPIPS}$  and Laplacian loss $\mathcal{L}_\text{lap}$ on the UV-space U-Net deformations. The total loss function thus becomes:
\begin{equation}
    \mathcal{L} = \mathcal{L}_\text{rgb} + \mathcal{L}_\text{lap} +  \mathcal{L}_\text{v,deform} +  \mathcal{L}_\text{v,rot} + \mathcal{L}_\text{scaling} + \mathcal{L}_\text{position}
\end{equation}
We set $\lambda_\text{v,rot} = 0.005$ and $\lambda_\text{v,deform} = 0.4$.

\paragraph{Back-face culling.}
After optimization, we find that, without special treatment, some Gaussians at the back of the head are not properly occluded by the front of the avatar. 
To mitigate artifacts from this effect, we cull all Gaussians that are attached to triangles that never face any camera during initialization. 
During training, we drop out all Gaussians that are attached to triangles that do not face the training view with a probability of 20\%. Finally, we set the initial number of Gaussians per triangle to one. 

\section{Evaluation Details}
\label{supp:evaluation_details}

\subsection{Self-reenactment}
\label{supp:sec_self_reenactment}

\paragraph{Dataset.}
We evaluate self-reenactment on sequences from the Nersemble~\cite{kirschstein2023nersemble} (forward-facing) and RenderMe-360~\cite{pan2024renderme} (360-degree) dataset. For each sequence, we obtain ground truth 3DMM parameters (FLAME~\cite{li2017learning}) using a multi-view face tracker \cite{taubner2024flowface} and all available views. Please see \cref{tab:supp_nersemble_self_sequences} for a full list of Nersemble sequences used for self-reenactment.

\begin{table}[t]
    \centering
    \caption{\textbf{Nersemble sequences for self-reenactment.} }
    \vspace{-3mm}
    \begin{tabular}{ l } 
        \toprule
        \textbf{sequence} \\ 
        \midrule
        \texttt{018-SEN-01-cramp\_small\_danger} \\
        \texttt{030-EMO-1-shout+laugh} \\
        \texttt{038-EXP-1-head} \\
        \texttt{085-SEN-09-frown\_events\_bad} \\
        \texttt{097-EXP-1-head} \\
        \texttt{124-FREE} \\
        \texttt{175-SEN-03-pluck\_bright\_rose} \\
        \texttt{226-EMO-4-disgust+happy} \\
        \texttt{227-SEN-08-clothes\_and\_lodging} \\
        \texttt{240-SEN-01-cramp\_small\_danger} \\
        \bottomrule
    \end{tabular}
    \label{tab:supp_nersemble_self_sequences}
\end{table}

\paragraph{Baseline implementation details.}
We implement and evaluate the baselines GAGAvatar~\cite{chu2024gagavatar}, Portrait4D-v2~\cite{deng2024portrait4dv2}, VOODOO3D~\cite{voodoo3d}, VOODOO XP~\cite{tran2024voodooxp}, HunyuanPortrait~\cite{xu2025hunyuanportrait}, CAP4D and CAP4D’s morphable multi-view diffusion model (MMDM)~\cite{taubner2024cap4d}. We provide the ground truth 3DMM parameters to all methods that can leverage 3DMM animations (\method and CAP4D), and use the corresponding ground truth videos and their monocular face tracker for the remaining methods (GAGAvatar, Portrait4D-v2, and VOODOO3D). These face trackers cannot process back-facing views, hence cannot be considered in our 360-degree evaluations. 

For our method, we pad the ground-truth 3DMM animations to 89 frames and use only 70 frames for evaluation or 4D reconstruction. 
The original CAP4D~\cite{taubner2024cap4d} implementation produces avatars that cover a view range of 55 and 20 degrees in azimuth and elevation respectively. To cover the higher view range of our evaluation set, we modify the generated view range of CAP4D to cover the same view range as our method: 60 degree azimuth and 30 degree elevation range.
As an additional 4D 360-degree baseline, we combine the portrait video generation model Follow-Your-Emoji (FYE)~\cite{ma2024followyouremoji} with the static multi-view generation model PanoHead~\cite{an2023panohead}. First, we use FYE to generate a reference video in a monocular setting using the reference image and the ground truth video from the reference view. Then, we extend this monocular video to multi-view by running PanoHead on each frame and using their resulting 3D representation to render the evaluation camera views. For this, we use the ground truth extrinsics for the novel view generation and the ground truth 3DMM to crop each video to the head. 

\paragraph{Metrics.}

We evaluate each method across the following metrics: photometric accuracy (PSNR, SSIM, LPIPS), temporal consistency (JOD)~\cite{mantiuk2021jod}, identity preservation (CSIM)~\cite{deng2022arcface}, and 3D consistency (RE@LG).
For all metrics, we center crop images around the projected vertices of the ground truth head model. For PSNR, SSIM, LPIPS and JOD, we mask both ground truth and predicted image using the background mask detected on the ground truth image using a background matting model~\cite{lin2021robustvideomatting}. 
To assess 3D consistency, we compute the reprojection error RE@LG following the implementation of Kant et al.~\shortcite{kant2025pippo}. First, we select four camera pairs per sequence based on angular proximity. Then, for each evaluation timestep, we detect DISK~\cite{tyszkiewicz2020disk} keypoints in all generated views. Then, we use LightGlue~\cite{lindenberger2023lightgluelocalfeaturematching} to match keypoints between camera pairs. Each matched keypoint is triangulated using known camera poses and intrinsics and projected to both views. The error between the reprojected triangulated point and the 2D keypoint positions is averaged across all matches, frames, and timesteps for this metric. 
To assess expression transfer accuracy, we include the average keypoint distance (AKD) measured using facial landmarks predicted from a keypoint detector \cite{bulat2017fan}, and the average expression distance (AED) predicted using DECA~\cite{deca}.

\paragraph{Additional results.}
We provide additional results on the self-reenactment task (more metrics and baselines) and report them in \cref{tab:supp_additional_self_reenactment}. Please find the complete collection of generated and evaluated sequences our project page. 

\begin{table}[]
    \centering
    \caption{\textbf{Additional self-reenactment results on the Nersemble dataset.}}
    \vspace{-3mm}
    \resizebox{\columnwidth}{!}{%
    \begin{tabular}{ l|c c c c c c c c} 
        \toprule
        Method & PSNR$\uparrow$ & SSIM$\uparrow$ & LPIPS$\downarrow$ & JOD$\uparrow$ & RE@LG$\downarrow$ & CSIM$\uparrow$ & AKD$\downarrow$ & AED$\downarrow$ \\ 
        \midrule
        HunyuanPortrait&18.04 & 0.671 & 0.380 & 4.02 & 4.32  & 0.641 & 8.20 & 1.11  \\   
        GAGAvatar     & 20.01 & 0.721 & 0.353 & 4.73 & 2.25  & 0.656 & 9.28 & 1.03  \\
        Portrait4D-v2 & 17.05 & 0.662 & 0.392 & 3.67 & 3.75  & 0.659 & 12.3 & 1.21 \\  
        VOODOO3D      & 20.09 & 0.692 & 0.349 & 4.86 & 1.84  & 0.517 & 7.34 & 1.27  \\  
        VOODOOXP      & 18.54 & 0.692 & 0.361 & 4.10 & 2.56  & 0.643 & 9.18 & 1.12  \\   
        CAP4D (MMDM)  & 22.17 & 0.768 & \best{0.280} & 5.30 & 1.28  & \best{0.793} & 4.22 & 0.78 \\ 
        CAP4D (120)   & 21.56 & 0.778 & 0.302 & 5.49 & \best{0.634} & \second{0.755} & 4.51 & 0.78  \\   
        \midrule
        MVP4D (MMVDM) & \best{23.39} & 0.765 & 0.294 & \best{5.88} & 1.03  & 0.707 & \best{4.10} & \best{0.76} \\   
        MVP4D (120)   & \second{23.16} & \best{0.790} & \second{0.301} & \second{5.62} & 0.656 & 0.708 & \second{4.33} & \best{0.76} \\  
        \bottomrule
    \end{tabular}}
    \label{tab:supp_additional_self_reenactment}
\end{table}

\subsection{Cross-reenactment}
\label{supp:sec_cross_reenactment}

\begin{table}[t]
    \centering
    \caption{\textbf{Nersemble driving sequences for cross-reenactment.}} 
    \vspace{-3mm}
    \begin{tabular}{ l } 
        \toprule
        \textbf{sequence} \\ 
        \midrule
        \texttt{018\_EXP-2-eyes} \\
        \texttt{030\_EMO-1-shout+laugh} \\
        \texttt{038\_EXP-3-cheeks+nose} \\
        \texttt{085\_SEN-01-cramp\_small\_danger} \\
        \texttt{097\_EMO-1-shout+laugh} \\
        \texttt{124\_EMO-4-disgust+happy} \\
        \texttt{175\_EXP-3-cheeks+nose} \\
        \texttt{226\_EMO-2-surprise+fear} \\
        \texttt{227\_SEN-10-port\_strong\_smokey} \\
        \texttt{240\_EMO-3-angry+sad} \\
        \bottomrule
    \end{tabular}
    \label{tab:supp_nersemble_cross_sequences}
\end{table}

\paragraph{Dataset.}

We evaluate cross-reenactment using 10 reference images from the FFHQ dataset~\cite{karras2019ffhq}, each selected to contain challenging geometric structure such as long hair, head wear or glasses. Each subject is paired with a driving video from the Nersemble evaluation dataset~\cite{kirschstein2023nersemble}, comprising 8 sequences with extreme expressions and 2 with speech, each with 75 frames @ 24 fps. Please see \cref{tab:supp_nersemble_cross_sequences} for a full list of Nersemble sequences used as driving videos for cross-reenactment. 
Each driving sequence is processed with a multi-view face tracker \cite{taubner2024flowface} to obtain the driving 3DMM animation (for \method and CAP4D). Methods that are unable to use 3DMM animation signals (Portrait4D-v2 and GAGAvatar) receive the forward-facing video to drive their avatar. 
We render each reenacted sequence with a camera following an elliptical trajectory around the head, sweeping up to 60° in azimuth and 20° in elevation. We complete one rotation with a period of 12 seconds, repeating the animation sequences up to this total time.  

\paragraph{User study.}

We ask participants to compare our method with a baseline method based on the following criteria:
\begin{itemize}
    \item \textbf{Level of Detail:} \textit{(Visual Detail)} Look closely at the face and head. Which video shows sharper details like skin texture, hair strands, blemishes, or wrinkles? Videos that look blurry should be rated lower.
    \item \textbf{Expression Matching:} \textit{(Expression Transfer)} Which avatar's facial expressions more accurately match the expressions in the original (driving) video?
    \item \textbf{3D Realism:} \textit{(3D Structure)} As the face turns or moves, does it maintain a realistic 3D shape? Rate lower if the face looks distorted or changes unnaturally as it moves or when viewed from different angles.
    \item \textbf{Natural Movement:} \textit{(Motion Quality)} Pay attention to how the face moves. Does it feel natural and lifelike, or stiff and robotic? Please also consider the movement of wrinkles, lips, and the tongue.
    \item \textbf{Overall Preference:} Based on everything above, which video do you personally prefer? Choose the one that is more appealing overall.
\end{itemize}

The video pairs are shown in random order, and each participant is asked to choose either the left or right video for each criterion. We collected 6750 responses from 45 participants and conducted $\chi^2$-tests to evaluate statistical significance at the $p < 0.05$ level; all results were statistically significant.

\begin{figure*}[ht]
\includegraphics[width=\textwidth]{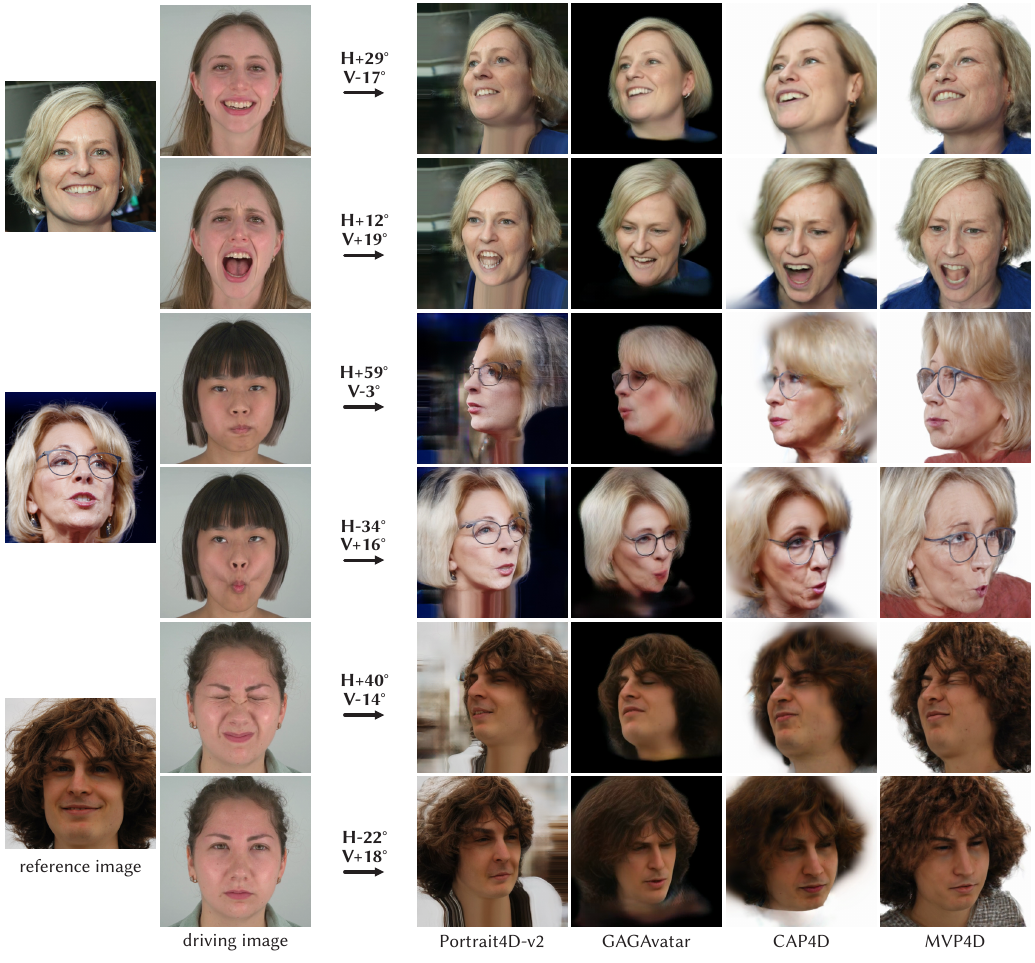}
\caption{\textbf{More cross-reenactment results.} Our method reconstructs challenging geometry such as hair better than previous methods, while also portraying more realistic expressions.} 
\label{fig:supp_cross_reenactment}
\end{figure*}

\paragraph{More results.}
We show more qualitative results in \cref{fig:supp_cross_reenactment}, and the complete collection of generated and evaluated sequences in the project page. 

\subsection{Multi-curriculum training}
\label{supp:sec_multi_training}

In the following, we provide information and experiments to motivate our multi-curricular training strategy. 

\paragraph{VRAM footprint during training}

In \cref{tab:supp_curriculum_vram}, we show the VRAM and iterations/day for each training stage (8x NVIDIA H100 with activation checkpointing to reduce VRAM). Full-resolution training (stage-3) requires significantly more memory and computation than other stages, motivating our curriculum. 

\begin{table}[t]
    \centering
    \caption{\textbf{Computational requirements during training.} VRAM footprint and speed (iterations per day) for each training stage. Full-resolution training requires significantly more memory and computation than early stages. }
    \vspace{-3mm}
    \begin{tabular}{ l|c c c } 
        \toprule
        stage & resolution & VRAM & iterations/day \\ 
        \midrule
        stage 1 & (1x49x256) & 29GB & 22k \\
        stage 2 & (1x49x512) & 41GB & 5k \\
        stage 3 & (4x49x512) & 50GB & 1k \\
        \bottomrule
    \end{tabular} 
    \label{tab:supp_curriculum_vram}
\end{table}

\paragraph{Performance for each stage.}

We evaluate performance after each training stage on our Nersemble self-reenactment benchmark (MMVDM only). Stage 1 performs significantly worse (no training at high resolution). Stage 2 performs slightly worse (no fine-tuning on many views/frames) (see \cref{tab:supp_staged_performance}). 

\begin{table}[t]
    \centering
    \caption{\textbf{Model performance after each training stage.}}  
    \vspace{-3mm}
    \begin{tabular}{ l|c c } 
        \toprule
        stage & PSNR$\uparrow$ & LPIPS$\downarrow$ \\ 
        \midrule
        stage 1 & 21.50 & 0.360 \\
        stage 2 & 22.90 & 0.303 \\
        stage 3 & 23.39 & 0.294 \\
        \bottomrule
    \end{tabular} 
    \label{tab:supp_staged_performance}
\end{table}

\paragraph{Ablation on multi-modal training curriculum.}

We conduct an ablation study of stages 1 and 2 at reduced spatial resolution (96x96 or 192x192) while keeping the same number of views/frames described in \cref{tab:training_stages}. 

Our proposed low-resolution training in stage 1 followed by stage 2’s higher-resolution fine-tuning (A, B) improves image quality vs. only stage 2 high-resolution training (C). We also observe an improvement in PSNR by limiting stage 1 training to mode 1 (A, proposed) instead of using modes 1-4 (B). \textbf{Settings A–C all use the same compute budget.} Results are shown in \cref{tab:supp_staged_ablation}. 

\begin{table}[t]
    \centering
    \caption{\textbf{Ablation on multi-modal training curriculum.} We compare different stage 1 and stage 2 combinations. The proposed curriculum (A) yields the best PSNR. }
    \vspace{-3mm}
    \resizebox{\columnwidth}{!}{%
    \begin{tabular}{ c| l | c c } 
        \toprule
        variation & description & PSNR$\uparrow$ & LPIPS$\downarrow$ \\ 
        \midrule
        A & \makecell[l]{(stage 1) 15k iters / 96x96 / mode 1 \\
                         (stage 2) 1k iters / 192x192 / mode 1-4} 
          & 21.6 & 0.392 \\ \addlinespace[0.5em]

        B & \makecell[l]{(stage 1) 15k iters / 96x96 / mode 1-4 \\
                         (stage 2) 1k iters / 192x192 / mode 1-4}
          & 21.2 & 0.386 \\ \addlinespace[0.5em]

        C & \makecell[l]{(stage 2 only) 5k iters / 192x192 / mode 1-4}
          & 20.5 & 0.385 \\ 
        \bottomrule
    \end{tabular} }
    \label{tab:supp_staged_ablation}
\end{table}

\subsection{Long video generation}
\label{supp:sec_long_video}

\paragraph{Computational requirements for longer videos.}

Below are the inference VRAM requirements for generating various video lengths (VxFxD).
\begin{itemize}
    \item 8x29x512: 27GB
    \item 8x49x512: 35GB
    \item 8x89x512: 42GB
\end{itemize}

We believe that our model can be distilled into more efficient feed-forward methods, just like \cite{deng2024portrait4dv2}. 

\paragraph{Temporal stability across temporal chunks.}

We assess generation of long videos (294 frames) in 6 temporal chunks (8 views, 49 frames, 512x512 resolution each) using generation mode 3 (\cref{fig:generation-modes}). Metrics are averaged over 4 RenderMe-360 sequences. As expected, iterated generation leads to some image quality deterioration (see \cref{tab:supp_temporal_chunks}).

\begin{table}[t]
    \centering
    \caption{\textbf{Long video model performance.} We report photometric quality across temporal chunks. Performance degrades with each iterated generation. }
    \vspace{-3mm}
    \begin{tabular}{ l|c c c } 
        \toprule
        frames & PSNR$\uparrow$ & LPIPS$\downarrow$ & JOD$\uparrow$ \\ 
        \midrule
        1-49    & 18.19 & 0.332 & 4.11 \\
        50-98   & 18.12 & 0.335 & 4.05 \\
        99-147  & 18.05 & 0.339 & 3.99 \\
        148-196 & 17.92 & 0.345 & 3.98 \\
        197-245 & 17.78 & 0.349 & 3.92 \\
        246-294 & 17.65 & 0.355 & 3.88 \\
        \bottomrule
    \end{tabular} 
    \label{tab:supp_temporal_chunks}
\end{table}

\end{document}